\documentclass[letterpaper, 10 pt, conference]{ieeeconf}  %

\IEEEoverridecommandlockouts                              %

\let\labelindent\relax

\usepackage{graphics} %
\usepackage{epsfig} %
\usepackage{times} %
\usepackage{amsmath} %
\usepackage{amssymb}  %
\usepackage{multirow}
\usepackage{pbox}
\usepackage{amsthm}
\usepackage{array} 
\usepackage{blindtext}
\usepackage{duckuments}
\usepackage{booktabs}
\usepackage[dvipsnames,table]{xcolor}
\usepackage{tabularx}
\usepackage{boldline, bbm}
\usepackage{arydshln} %
\usepackage{color}
\usepackage{enumitem}
\usepackage{multicol}
\usepackage{float}
\usepackage{pifont}
\usepackage{cite}
\usepackage{bbm}
\usepackage{graphicx}
\graphicspath{{./figures/}}
\usepackage{multicol}        %
\usepackage{algorithm}
\usepackage[font=footnotesize,labelsep=period]{caption}[2022/02/20]
\usepackage{subcaption}
\usepackage{algpseudocode}
\usepackage{makecell}
\usepackage[export]{adjustbox}
\makeatletter
\let\NAT@parse\undefined
\makeatother
\usepackage[
colorlinks,
linkcolor=blue,
citecolor=blue,
filecolor=red,
urlcolor=blue]{hyperref}
\usepackage{balance}

\newcommand{\PlannerName}{$\textbf{\textit{\texttt{PIPE}}}$}%
\newcommand{\etal}{\textit{et al.}~}

\newcommand{\xxnote}[3]{}
\ifx\hidenotes\undefined
  \renewcommand{\xxnote}[3]{\color{#2}{#1: #3}}
\fi

\title{\LARGE \bf
PIPE Planner: Pathwise Information Gain with \\ Map Predictions for Indoor Robot Exploration

}

\algrenewcommand\algorithmicrequire{\textbf{Input:}}
\algrenewcommand\algorithmicensure{\textbf{Output:}}

\author{Seungjae Baek$^{*1,2}$, Brady Moon$^{*2}$, Seungchan Kim$^{*2}$, Muqing Cao$^{2}$, \\ Cherie Ho$^{2}$, Sebastian Scherer$^{\dag2}$, Jeong hwan Jeon$^{\dag1}$
\thanks{*: Equal Contributions \hspace{1.5mm} $\dag$: Equal Advising}%
\thanks{This work was supported by the National Institute of Advanced Industrial Science and Technology (AIST), DSTA contract No. DST000EC124000205, NSF GRFP under Grant No. DGE1745016, a hardware grant from Nvidia, and by Institute of Information \& communications Technology Planning \& Evaluation (IITP) grants by the Korean government (MSIT) (No.RS-2022-00143911, AI Excellence Global Innovative Leader Education Program; No.RS-2020-II201336, Artificial Intelligence graduate school support (UNIST); and No.RS-2024-00341055).}
\thanks{$^{1}$ Authors are with Ulsan National Institute of Science and Technology, Ulsan, Republic of Korea. \tt{\{bsj970, jhjeon\}@unist.ac.kr}}
\thanks{$^{2}$ Authors are with the Robotics Institute, School of Computer Science at Carnegie Mellon University, Pittsburgh, PA, USA. \tt{\{bradym, seungch2, muqingc, cherieh, basti\}@andrew.cmu.edu}}
}

\begin{document}

\maketitle
\thispagestyle{empty}
\pagestyle{empty}

\begin{abstract}
Autonomous exploration in unknown environments requires estimating the information gain of an action to guide planning decisions. While prior approaches often compute information gain at discrete waypoints, pathwise integration offers a more comprehensive estimation but is often computationally challenging or infeasible and prone to overestimation. In this work, we propose the Pathwise Information Gain with Map Prediction for Exploration (\PlannerName) planner, which integrates cumulative sensor coverage along planned trajectories while leveraging map prediction to mitigate overestimation. To enable efficient pathwise coverage computation, we introduce a method to efficiently calculate the expected observation mask along the planned path, significantly reducing computational overhead. We validate \PlannerName~on real-world floorplan datasets, demonstrating its superior performance over state-of-the-art baselines. Our results highlight the benefits of integrating predictive mapping with pathwise information gain for efficient and informed exploration. Website: \href{https://pipe-planner.github.io}{pipe-planner.github.io}
\end{abstract}

\section{Introduction}
\label{sec:introduction}
Autonomous robots are often deployed to explore in unknown environments, where they must make decisions and take actions without knowing what future observations will reveal \cite{yamauchi1997frontier,umari2017autonomous, scherer2022resilient}. A key challenge of exploration is estimating the potential information gain from each action \cite{stachniss2005information}, which quantifies the value of the resulting sensor observations using metrics such as coverage \cite{bircher2016receding, delmerico2018comparison}, uncertainty reduction \cite{vallve2014dense}, or map quality. This challenge is particularly important in indoor exploration, where limited sensing range and complex layouts can lead to inefficient navigation without informed decision-making \cite{kim2023multi}.

A common approach to computing predicted information gain for an exploration action is to estimate the expected gain at possible next waypoints. For example, Bircher \etal \cite{bircher2016receding} iteratively selects viewpoints that maximize the volume of unmapped, visible voxels within the sensor's field of view. These methods, which evaluate information gain at specific waypoints, can be referred to as \textit{pointwise} information gain-based approaches. In contrast, a more comprehensive and accurate strategy is to adopt a \textit{pathwise} information gain metric, which optimizes planned paths based on total information gain accumulated along the robot’s trajectory. This better matches reality, where most robotic sensors continuously gather information, not merely at waypoints. 
Prior works have explored different aspects of pathwise information gain. In informative path planning, Binney \etal \cite{binney2013optimizing} proposed a broader formulation that accounts for information gain not only at individual graph nodes but also sensor measurements along the edges, such as temperature or depth, rather than modeling broader sensor coverage along the path. More specialized approaches, such as \cite{liu2022estimated}, focus on modeling sensor coverage along a path. 

\begin{figure}[t]
    \centering
\includegraphics[width=\linewidth]{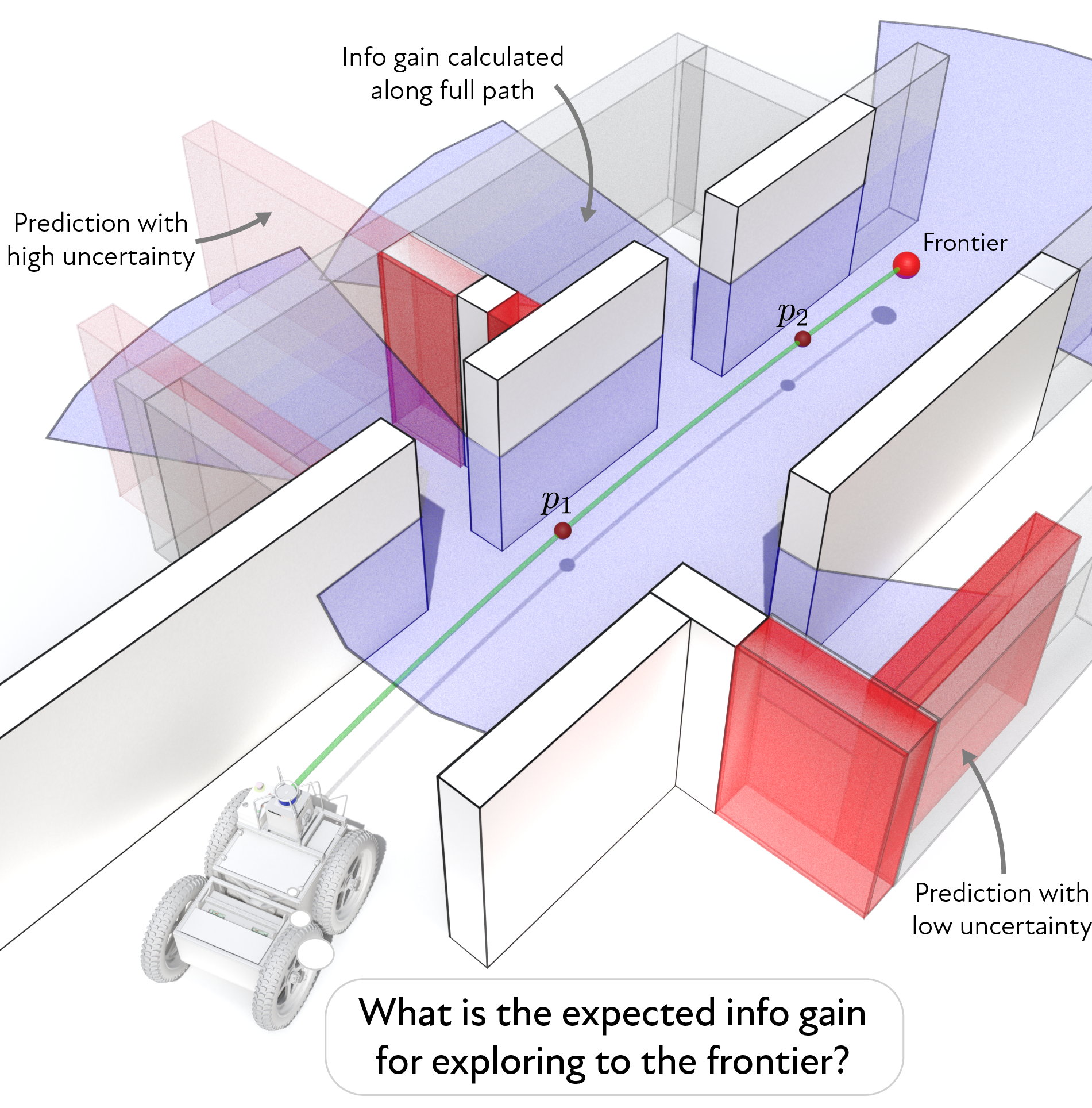}    
\caption{Pathwise exploration often suffers from overestimating information gain. Our planner, \PlannerName, solves this by using a map predictor to accurately calculate the expected information gain over the entire path, reasoning about both visibility and map quality.}
\label{intro-figure}
\vspace{-0.60cm}
\end{figure}

In this work, we focus on pathwise sensor coverage, a subset of pathwise information gain that considers the sensor’s ability to observe and map the environment along a planned trajectory. Pathwise integration of sensor coverage presents several challenges. First, it is often computationally expensive; for example, estimating camera or LiDAR coverage at every point along a path requires significant computation. Finding an efficient integration method remains a non-trivial problem. Second, information gain can be overestimated, especially when undiscovered obstacles or occlusions exist beyond the robot’s current field of view. Third, these estimation errors can accumulate over the trajectory, leading to compounding inaccuracies and suboptimal planning.

To address these challenges, we first focus on efficiently computing integrated information gain along a planned trajectory. We propose an approach that effectively estimates sensor coverage along the path by insightfully leveraging computational geometry, significantly reducing computational overhead. Second, to mitigate overestimation and error accumulation, we integrate map prediction into the robot’s raycasting for sensor coverage estimation. Prior research \cite{8793500, shrestha2019learned, georgakis2022uncertainty, ho2024mapex} has shown that deep learning models can predict maps beyond observed areas, enhancing exploration in various ways. We specifically use such predictive models to refine sensor coverage estimates by generating plausible map predictions ahead of the robot. To the best of our knowledge, our work is the first to integrate map prediction-based exploration with a pathwise information gain metric, enabling more efficient exploration planning.%

Combining these two aspects, we propose \textbf{P}athwise \textbf{I}nformation Gain with Map \textbf{P}rediction for \textbf{E}xploration (\PlannerName) planner. \PlannerName~generates predictions of map beyond observed areas, computes cumulative sensor coverage along planned trajectories, and leverages predictive uncertainties as an information gain metric for planning. We validate \PlannerName~on a real-world floorplan dataset, and show that it outperforms SOTA baselines across multiple evaluations. 

The contributions of this work are as follows:
\begin{itemize}
    \item We propose an indoor robot exploration planner, \PlannerName, that estimates information gain using path-integrated cumulative sensor coverage and map predictions.
    \item We propose a fast method for estimating expected pathwise info gain by leveraging computational geometry.
    \item We show that \PlannerName~outperforms state-of-the-art indoor exploration baselines on a real-world floorplan dataset in both budget-constrained and full exploration.
\end{itemize}

\section{Related Work}
\label{sec:related_work}
\subsection{Robot Exploration}
Robot exploration is the process of sensing an environment and finding feasible paths to construct a map. Early approaches, such as frontier-based exploration \cite{yamauchi1997frontier}, incrementally explore the environment by sequentially visiting frontiers---the boundary edges between observed and unobserved spaces. Other methods, known as information gain-based approaches, estimate information gain and select actions that maximize it. Various information gain metrics have been proposed, including maximizing 2D sensor coverage \cite{gonzalez2002navigation, deng2020robotic}, optimizing for 3D volumetric information gain \cite{bircher2016receding, delmerico2018comparison, corah2021volumetric}, and reducing entropy or uncertainty \cite{vallve2014dense}. In this work, our information gain metric is a combination of sensor coverage and uncertainty reduction.

\subsection{Path Planning with Pathwise Information Gain}
In path planning, reasoning about a robot's viewpoint is a common approach. Estimating the amount of information a robot can observe within its sensor's field of view at a given point has been widely studied in various areas, such as data gathering \cite{popovic2017online, moon2025iatigris}, reconstruction \cite{maboudi2023review}, inspection \cite{bircher2015structural}, and exploration \cite{gonzalez2002navigation, bircher2016receding, charrow2015information, zhang2024falcon}. Beyond viewpoint-based reasoning, some works have explored more densely sampled points along planned trajectories in the context of path planning. Binney \etal \cite{binney2013optimizing} proposed variants of informative path planning algorithms that leverage edge-based samples, taking dense points along edges into account. In contrast with sampling points along a path, Moon \etal \cite{moon2022tigris} approximates the information gain along the path using an efficient lower bound. The closest work to ours is Liu \etal \cite{liu2022estimated}, which aims to estimate path information gain for exploration by measuring sensor coverage along the trajectory from the current frontier to the next. However, their approach approximates cumulative sensor coverage using a parallelogram-based estimation, which doesn't capture accurate information gain within complex layouts like indoor environments. Our work differs from previous approaches by providing a precise estimation of the information gain a robot can obtain along its path using raycast modeling. Additionally, we introduce a technique for efficiently computing the sensor coverage area, enabling more accurate and computationally feasible path information gain estimates for exploration planning.

\subsection{Map Prediction for Exploration}
Recent studies in exploration go beyond using only the observed area for path planning; instead, they leverage broader map predictions to improve planning. This technique, commonly used in indoor robot exploration, involves training deep learning models on offline datasets to generate plausible map predictions derived from the observed portion of a top-down occupancy map. These approaches guide robots to maximize information gain by leveraging the generated map predictions. For instance, \cite{georgakis2022uncertainty} directs the robot toward more uncertain areas by summing variances along the trajectory, but it does not account for sensor coverage. In contrast, \cite{shrestha2019learned} guides the robot to regions where predicted sensor coverage is expected to be higher. A recent work \cite{ho2024mapex} enables the robot to probabilistically reason about both sensor coverage and uncertainty simultaneously, choosing frontiers with a combination of high uncertainty and high coverage.

Our work also utilizes map prediction as a crucial component for exploration, helping to avoid overestimation when computing pathwise information gain. Without predicted maps, a robot planning a path beyond its observed area risks overestimating the amount of information its sensors could acquire. By assuming a plausible predicted map, we establish a better estimate of the environment and provide a reasonable bound on the expected information gain along the path.

\begin{figure*}[t]
    \centering
\includegraphics[trim={0cm 6.0cm 0cm 0.55cm},clip,width=\linewidth]{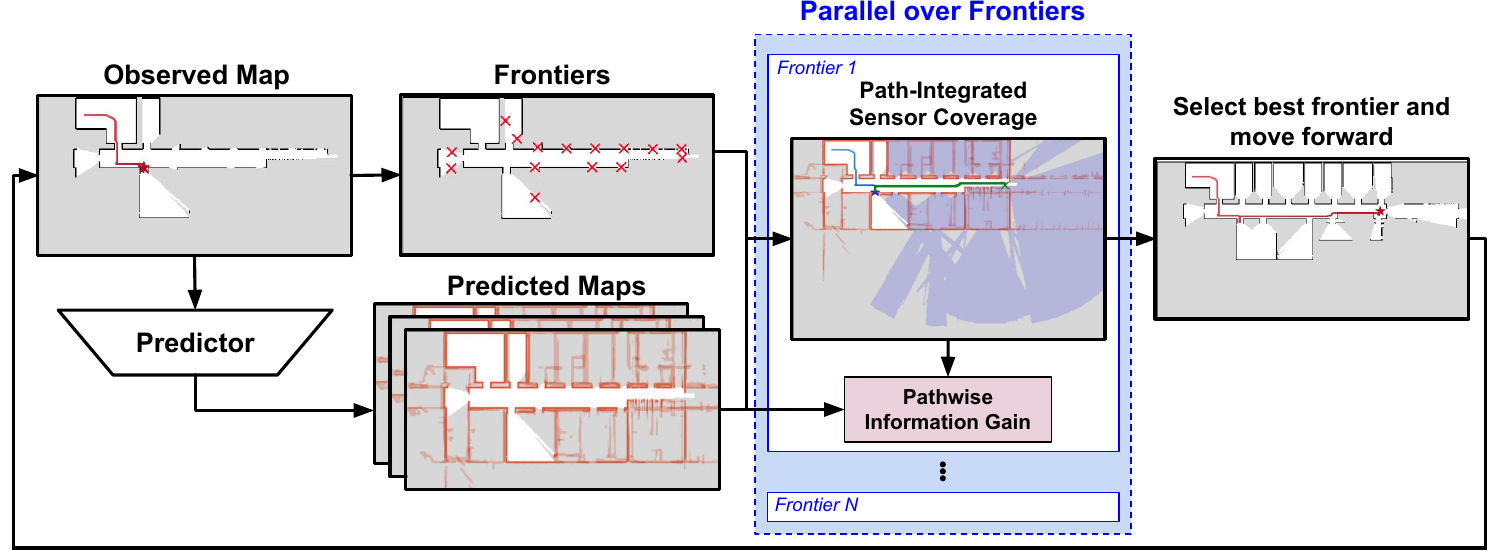}   
\caption{Overview of \PlannerName~planner pipeline. The robot extracts frontiers from its observed occupancy grid map and generates predicted maps. For each frontier, \PlannerName~estimates path-integrated sensor coverage and generates a visibility mask for the full path. It computes pathwise information gain by summing the pixelwise variance of predictions within the visibility mask. The robot then selects the frontier with the highest information gain for exploration planning.}
\label{pipeline}
\vspace{-0.4cm}
\end{figure*}

\section{Problem Statement}
\label{sec:problem_statement}

We define the indoor robot exploration problem as a robot navigating a 2D indoor environment $\mathcal{E} \subset \mathcal{R}^2$ to build a map of the environment. The robot's state at time $t$ is given by $\mathbf{x}_t = [x_t, y_t]$, where $x_t, y_t \in \mathbb{Z}$. It perceives its surroundings using a noise-free 2D LiDAR sensor with a range of $\lambda$ and updates an (observed) occupancy grid map $O_t$, where each cell is classified as free, occupied, or unknown. The sensor collects $l$ evenly distributed samples per scan over a full $360^\circ$ field of view. At each time step, the robot moves $\Delta$ by selecting an action $a \in \mathcal{A}$, where $\mathcal{A}$ is a movement on an 8-connected grid. The objective of exploration is to determine a feasible path $\{\mathbf{x}_0, \mathbf{x}_1, \ldots, \mathbf{x}_T\}$ that constructs an accurate map representation within a fixed time $T$.

\section{Approach}
\label{sec:approach}

In this section, we introduce the approach for \PlannerName~planner, as summarized in Fig.~\ref{pipeline}. We begin by discussing exploration frameworks in Sec.~\ref{IV-A} and introducing the concept of sensor coverage and visibility masks in Sec.~\ref{IV-B}. In Sec.~\ref{IV-C}, we propose a path-integrated sensor coverage approach for computing information gain, followed by our methods to accelerate computation and optimize performance in Sec.~\ref{IV-D}. Finally, in Sec.~\ref{IV-E}, we explain how the pathwise information gain metric is integrated into the exploration planner and propose the \PlannerName~planner.

\subsection{Preliminary I: Exploration Framework}
\label{IV-A}
In this work, we use classical frontier-based exploration \cite{yamauchi1997frontier} as our planning framework. In frontier-based approaches, the robot identifies boundaries between known and unknown regions and selects the nearest frontier to explore. Building on this, we incorporate an information gain metric to select the next frontier to travel that maximizes expected information gain, following the ``next-best-view" paradigm \cite{gonzalez2002navigation, visser2008beyond, bircher2016receding,  vasquez2017view,delmerico2018comparison}. 

To quantify the expected information gain, we first generate a predicted map $M_t$ from the observed occupancy grid $O_t$ using a pretrained image inpainting network \cite{suvorov2022resolution}, similar to other prior map prediction-based exploration methods \cite{shrestha2019learned, georgakis2022uncertainty, ho2024mapex}. While frontiers are extracted from $O_t$, the information gain for traveling to the frontier is evaluated using both $O_t$ and $M_t$ by choosing actions that will reduce the uncertainty in $M_t$. Following \cite{georgakis2022uncertainty, ho2024mapex}, we estimate prediction uncertainty map $U_t$ by computing the pixelwise variance across an ensemble of predicted maps. The robot then selects the frontier that will result in observing the highest total amount of uncertainty relative to the path cost. In contrast with previous works, our work selects not the next best frontier observation but rather the holistic next best frontier by including the path to the frontier.

\subsection{Preliminary II: Sensor Coverage and Visibility Masks}
\label{IV-B}
To estimate the LiDAR sensor coverage that a robot can obtain at a state $\mathbf{x}$, it must perform a raycast by emitting virtual rays from $\mathbf{x}$. For example, when calculating the sensor coverage at a frontier $f$ in the 2D NBV-based methods \cite{gonzalez2002navigation, visser2008beyond}, we perform a raycast from $f$ with a range of $\lambda$. The ray stops when it hits an occupied cell in $O_t$, while in free space, it extends until it reaches the range limit $\lambda$. 
\begin{equation}
    \{v_{1:l}\} \leftarrow \textsc{Raycast} (f, \lambda, O_t) 
\end{equation}
Here, $\{v_{1:l}\}$ are the vertices that each ray hits at the map, and $l$ is the number of laser samples. 

As we leverage map predictions, we can perform raycast on predicted map $M_t$ instead of $O_t$, 
\begin{equation}
    \{v_{1:l}\} \leftarrow \textsc{Raycast} (f, \lambda, M_t) 
\end{equation}
While we can apply the same raycasting method to the predicted map $M_t$, where the ray stops upon hitting a predicted occupied cell, prior work \cite{ho2024mapex} has shown that this approach can underestimate sensor coverage, particularly when erroneous predictions falsely classify free space as occupied. To address this, we adopt the probabilistic raycasting method proposed in \cite{ho2024mapex}. Instead of terminating immediately upon encountering a predicted occupied cell, the ray starts with an initial value $\delta=0$, and incrementally accumulates the values of predicted cells it traverses. Once $\delta$ reaches a predefined threshold $\epsilon$, the raycasting stops. This probabilistic approach allows for tuning of the sensor coverage estimation by adjusting $\epsilon$, balancing between overestimation and underestimation. Following prior work, we set $\epsilon=0.8$. 

From the set of $l$ vertices generated from the raycast, we can draw a polygon $\mathcal{G}$, and use a flood fill algorithm to generate a visibility mask $\nu$, which is an estimate sensor coverage at the frontier point $f$. 
\begin{equation}
    \begin{aligned}
        \mathcal{G} &\leftarrow \textsc{DrawPolygon}(\{v_{1:l}\}) \\
        \nu &\leftarrow \textsc{FloodFill} (\mathcal{G})
    \end{aligned}
\end{equation}

\begin{figure}[t]
    \centering
    \includegraphics[trim={12cm 2.5cm 14.2cm 5.5cm},clip,width=\linewidth]{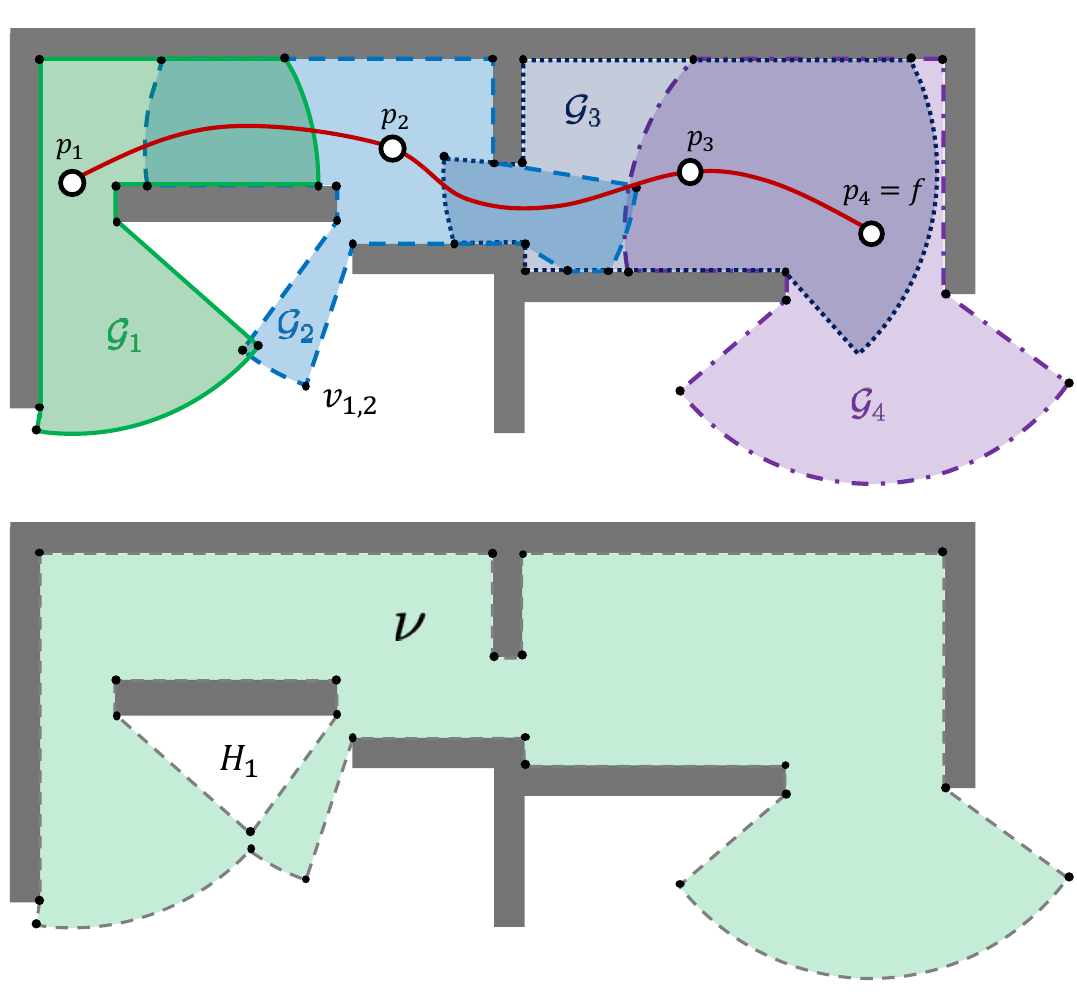}
    \caption{A visualization of the path visibility mask computation. Individual polygons are first generated using raycasting. These polygons are then merged into a single polygon, including any holes, and a flood-fill algorithm is applied to determine the final visibility mask.}
    \label{fig:floodfilling}
    \vspace{-0.2cm}
\end{figure}

\subsection{Path-Integrated Sensor Coverage for Information Gain}
\label{IV-C}
The core component of our \PlannerName~planner is to evaluate the information gain of each frontier point $f \in \mathcal{F}$ (where $\mathcal{F}$ is a set of frontiers) by integrating the cumulative information gain along the path from the robot's current pose $\mathbf{x}_t$ to $f_i$, rather than estimating sensor coverage solely at $f_i$. 

For each frontier $f_i \in \mathcal{F}$, we generate an A* path $P_i$ connecting the robot's current pose $\mathbf{x}_t$ to $f_i$. Along $P_i$, we sample each point $p_j$ to perform a raycast with a range of $\lambda$ on the predicted maps $M_t$. These sampled points along $P_i$ are every $\Delta$, which is the same distance for each robot time step and update to the observed occupancy grid map $O_t$. 

A simple approach to calculating the total visibility mask along the path is to first generate the visibility masks $\nu_j$ for each $p_j$:  
\begin{equation}
    \begin{aligned}
        \{v_{1:l}\}_j &\leftarrow \textsc{Raycast}(p_j, \lambda, M_t) \\
        \mathcal{G}_j &\leftarrow \textsc{DrawPolygon}(\{v_{1:l}\}_j) \\
        \nu_j &\leftarrow \textsc{FloodFill}(\mathcal{G}_j)
    \end{aligned}   
\end{equation}
The mask $\nu_j$ represents the sensor coverage the robot can obtain at each point $p_j$ along the path to the frontier $f_i$. The path-integrated sensor coverage, $\nu$, is then obtained by computing the union of each $\nu_j$ along the path. 
However, this straightforward approach to calculate pathwise information gain introduces significant computational overhead.

\subsection{Optimization of Calculating Information Gain}
\label{IV-D}
\begin{figure}[t]
    \centering
\includegraphics[width=\columnwidth]{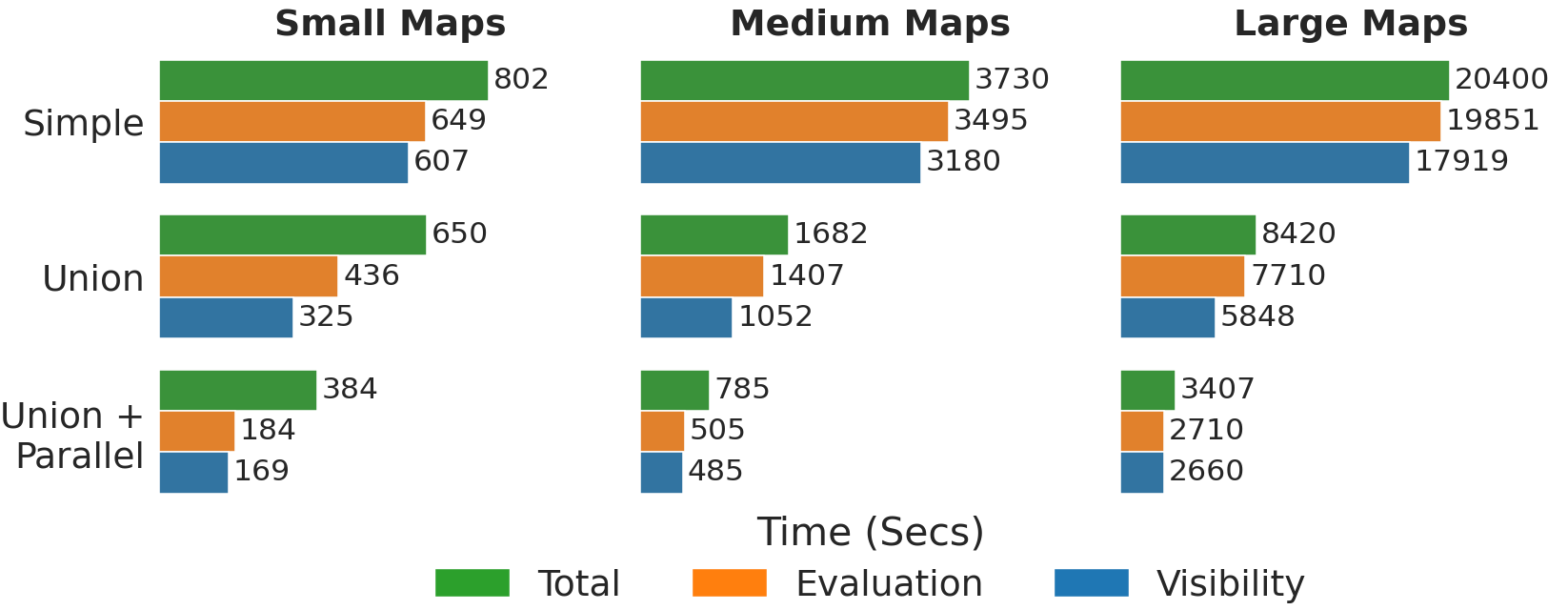}
    \caption{Computation time comparison of our optimized approach (union + parallel) across map sizes, with the greatest reduction for large maps.}
    \label{fig:bargraph}
\end{figure}

To identify the most computationally demanding components of our naive planner implementation, we conducted a simple experiment. We ran our planner on two small, two medium, and two large maps from the KTH dataset \cite{aydemir2012can} (which will be further detailed in Sec.~\ref{sec:experimental-setup}), each with two different starting positions (one from a corner and one from the center), measuring the computation time. Computation time comparisons were conducted on a desktop computer equipped with an \textit{Intel Core i5-10400F} CPU (2.90GHz), an \textit{NVIDIA GeForce RTX 3060} GPU, and 16GB of RAM.
The results are shown in Fig. \ref{fig:bargraph}. As expected, we found that in the simple approach, frontier evaluations accounted for 80.9\%, 93.7\%, and 97.3\% of the total computation time in small, medium, and large maps, respectively. The increasing proportion in larger maps is likely due to the greater number of frontiers and longer path lengths. Among frontier evaluations, visibility mask generation was the most time-consuming step (blue bars in the figure), with repeated flood fill operations being the primary bottleneck.

\begin{algorithm}[t]
\caption{PathVisibilityMask}\label{alg:polygon} 
\begin{algorithmic}[1]
\Require Path $P_i$ (from the robot's current pose $\mathbf{x}_t$ to the candidate frontier $f_i$), map $M$, LiDAR range $\lambda$ %
\State $\mathbb{G} = \{\}$ \Comment{Initialize}
\State \textbf{for} each point $p_j \in P_i$:
\State \hspace{1.5mm} $\triangleright$ \textbf{Probabilistic Raycast and Polygon Construction:}
\State \hspace{1.5mm} $\{v_{1:l}\}_j \leftarrow \textsc{Raycast}(p_j, \lambda, M)$ \Comment{Vertices}
\State \hspace{1.5mm} $\mathcal{G}_j \leftarrow \textsc{DrawPolygon}(\{v_{1:l}\}_j)$ \Comment{Polygon}
\State \hspace{1.5mm} $\mathbb{G} \leftarrow \mathbb{G} \cup \{\mathcal{G}_j\}$
\State $\triangleright$ \textbf{Merge Polygons and Generate Visibility Mask:}
\State $\mathcal{G}_{union} \leftarrow \bigcup_{\forall j}^{}  \mathcal{G}_j$
\State $S \leftarrow \bigcup_{\forall j} (\mathcal{G}_j \:\triangle\:\mathcal{G}_{union} )$
\State $H \leftarrow S \: \backslash \: \mathcal{G}_{union}$
\State $\nu \leftarrow$ \textsc{FloodFill} $(\mathcal{G}_{union},H)$
\Ensure Visibility Mask $\nu$
\end{algorithmic}
\end{algorithm}

To mitigate the prohibitive computational overhead, our implementation avoids applying a separate flood-fill to each raycast polygon $\mathcal{G}_j$. Instead, we first compute the union of all polygons to form a single composite shape, $\mathcal{G}_{union}$. By deferring the expensive flood-fill to a single, final step on this merged polygon, the overall computational cost is substantially reduced.

Since most paths contain multiple, typically nonconvex raycasted polygons, their union often creates holes $H$, or trapped regions. To identify these holes and account for them in the flood fill operation, we follow these steps: first, compute the symmetric difference $S_j$ by performing symmetric difference operation $\triangle$ between each polygon $\mathcal{G}_j$ and the polygon union ($S \leftarrow \bigcup_{\forall j} \:(\mathcal{G}_j \:\triangle\:\mathcal{G}_{union} )$). Then extract the parts of $S$ that are not contained in the polygon union ($H \leftarrow S \: \backslash \: \mathcal{G}_{union}$). 
We then compute the final visibility mask $\nu$ using the polygon union and holes ($\nu \leftarrow \textsc{FloodFill}(\mathcal{G}_{union}, H)$). The full pseudocode for this process is shown in Alg.~\ref{alg:polygon}, with a visualization in Fig.~\ref{fig:floodfilling}.

\begin{algorithm}[t]
\caption{\PlannerName~Planner}\label{alg:pipe} 
\begin{algorithmic}[1]
\Require Robot's start pose $\mathbf{x}_0$, Time budget $T$
\State \textbf{while}  $t \leq T$:
\State \hspace{1.7mm} $\triangleright$ \textbf{Sense and Update Map:}
\State \hspace{1.7mm} $O_t \leftarrow$ \textsc{Raycast}($\mathbf{x}_t, \lambda, O_t$)
\State \hspace{1.7mm} \textbf{if} robot needs to select a new waypoint:
\State \hspace{3.4mm} $\triangleright$ \textbf{Extract Frontiers and Predictions:}
\State \hspace{3.4mm} $\mathcal{F} \leftarrow$ \textsc{ExtractFrontiers}($O_t$)
\State \hspace{3.4mm} $M_t \leftarrow$ \textsc{GeneratePredictions}($O_t$) \Comment{Ensemble}
\State \hspace{3.4mm} $U_t \leftarrow \textsc{ComputeVariance}(M_t)$ %
\State \hspace{3.4mm} $\triangleright$ \textbf{Evaluate Frontiers:}
\State \hspace{3.4mm} \textbf{for} frontier $f$ in $\mathcal{F}$:
\State \hspace{5.5mm} $P \leftarrow$ \textsc{AStarPath}($\mathbf{x}_t,f$)
\State \hspace{5.5mm} $\nu \leftarrow \textsc{PathVisibilityMask}(P, M_t, \lambda)$ \Comment{Alg.~\ref{alg:polygon}}
\State \hspace{5.5mm} $I \leftarrow \sum_{(x_k,y_k)\in \nu} U_t(x_k,y_k) $ 
\State \hspace{5.5mm} $f.\text{score} \leftarrow I/|P|$ \Comment{Normalize by path length}
\State \hspace{3.4mm} $f_{\text{max}} \leftarrow \underset{f\in \mathcal{F}}{\arg \max}f.\text{score}$
\State \hspace{1.7mm} $\triangleright$ \textbf{Plan Local Actions:}
\State \hspace{1.7mm} $a_t \leftarrow$ \textsc{AStarPath}($\mathbf{x}_t, f_{\max}$)
\State \hspace{1.7mm} $\mathbf{x}_{t+1} \leftarrow \textsc{Transition}(\mathbf{x}_t, a_t)$
\State \hspace{1.7mm} $t \leftarrow t + 1$
\Ensure Final observed map $O_T$, final predicted map $M_T$
\end{algorithmic}
\end{algorithm}

Additionally, to take advantage of computational parallelism, we used Python’s multiprocessing package to distribute the computation across multiple CPU cores, with each process handling a separate frontier. In an ideal scenario, this approach can reduce computation time by a factor approximately equal to the number of available CPU cores, assuming efficient implementation and minimal overhead.

The impact of these optimizations, shown in Fig.~\ref{fig:bargraph}, is most pronounced in the large maps, with a $83.3\%$ reduction in computation time.

\subsection{\PlannerName~Planner}
\label{IV-E}
Lastly, we outline the overall procedure of \PlannerName~planner. When the robot needs to select a new waypoint, it first extracts a set of candidate frontier points, $\mathcal{F}$, from the current occupancy grid $O_t$, and generates predictions $M_t$ and uncertainty map $U_t$. For each frontier $f\in \mathcal{F}$, it generates an A* path, $P$, from current pose $\mathbf{x}_t$ to the frontier $f$, and then applies Alg.~\ref{alg:polygon} to compute the pathwise sensor coverage, yielding the visibility mask $\nu$. The information gain, $I$, is then calculated by summing the uncertainties within the visibility mask and normalizing by the length of the path $P$. The frontier with the highest score is selected as the next waypoint until it has been reached, at which a new waypoint is selected through the same process described as above. The complete procedure of \PlannerName~planner is presented in Alg.~\ref{alg:pipe}.

\setlength{\fboxrule}{0.8pt} %
\setlength{\fboxsep}{0pt}   %

\section{Experiments}
\label{sec:experiments}
In this section, we describe the experimental setup, baselines, and evaluation metrics, followed by the experimental results. We provide both qualitative and quantitative analyses comparing our \PlannerName~planner against the baselines.  

\begin{figure}[t]
    \centering
\includegraphics[trim=2.9cm 1.1cm 11.2cm 0.45cm, clip, width=\linewidth]{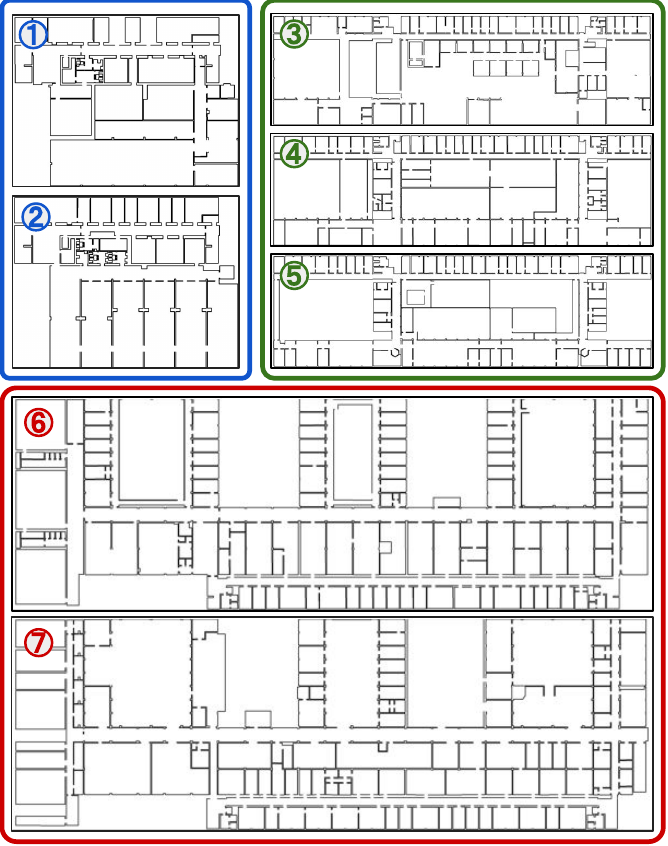} 
\caption{The seven indoor maps used to evaluate our method. Small, medium, and large maps are shown in blue, green, and red, respectively. The maps have various topologies and are not shown to scale relative to each other. }
\label{fig:kth-test-environments}
\end{figure}

\begin{figure*}[t]
    \centering
\includegraphics[width=\linewidth]{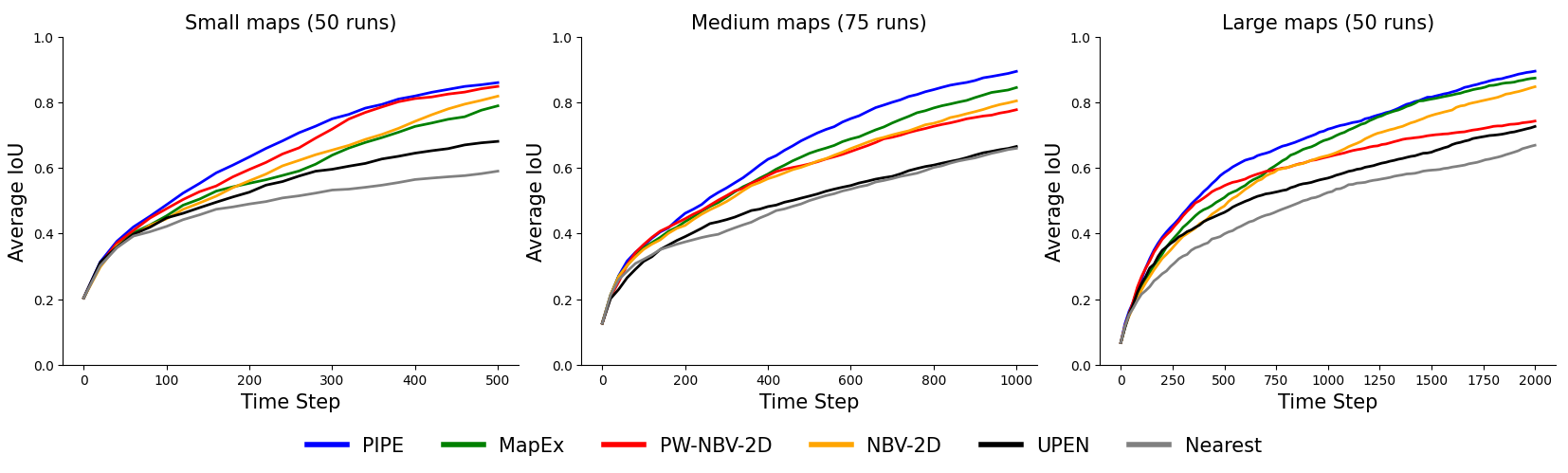}    
\caption{Quantitative comparison of methods for each map size shows \PlannerName~outperforms all others in most time steps, with the second-best method varying by map size. Prediction-based methods excel on larger maps, while pathwise NBV-2D performance declines as map size increases.}
\label{fig:result}
\end{figure*}

\subsection{Experimental Setup}
\label{sec:experimental-setup}
We use a custom-built simulator based on the KTH floorplan dataset \cite{aydemir2012can}, which includes over 100 campus floorplans represented as XML files containing wall and door locations for each room. To ensure accurate raycasting, we corrected any inaccuracies in the maps that could lead to false projections into invalid spaces. These corrected maps serve as the simulator's ground truth, while the planner itself utilizes the LaMa model \cite{suvorov2022resolution} to generate map predictions from online observations. The maps were downsampled to a resolution of 10 pixels per meter. 

From the dataset, we selected seven floorplans, categorized into three sizes: two small ($65\,\text{m} \times 85\,\text{m}$), three medium ($60\,\text{m} \times 205\,\text{m}$), and two large maps ($88\,\text{m} \times 265\,\text{m}$). Fig.~\ref{fig:kth-test-environments} visualizes these environments. For each floorplan, the robot starts from 25 distinct initial locations. In total, we conducted 175 experiments to evaluate performance.

\begin{table}[t]
\centering
\renewcommand{\arraystretch}{1.2}
\begin{tabular}{lccc} %
\toprule
\textbf{Method} & \textbf{Small Maps} & \textbf{Medium Maps} & \textbf{Large Maps} \\
\midrule
\PlannerName~(ours) & \textbf{327.00} & \textbf{648.35} & \textbf{1338.16} \\
MapEx \cite{ho2024mapex} & 290.63 & 607.05 & 1278.39 \\
PW-NBV-2D & 316.81 & 587.59 & 1184.69 \\
NBV-2D \cite{visser2008beyond} & 295.68 & 582.70 & 1209.37 \\
UPEN \cite{georgakis2022uncertainty} & 270.04 & 494.08 & 1084.57 \\
Nearest \cite{yamauchi1997frontier} & 244.80 & 482.79 & 971.30 \\
\bottomrule
\end{tabular}
\caption{Area Under the Curve (AUC) of IoU for each method. Higher is better. \PlannerName~outperforms other methods in all maps.}
\label{AUC}
\vspace{-0.25cm}
\end{table}

\subsection{Baselines}
We select five baselines for comparison, a conventional nearest frontier-based method, two for observed map based method, and other two for map prediction based exploration methods.
\begin{itemize}
    \item Nearest \cite{yamauchi1997frontier}: A classical exploration method that selects frontiers based on the shortest Euclidean distance. 
    \item NBV-2D \cite{gonzalez2002navigation, visser2008beyond}: A ``Next-best-view'' strategy where, for each frontier, the robot performs virtual raycasting on the observed map $O_t$, computes sensor coverage, and normalize it by the Euclidean distance. \textit{(Pointwise, No Prediction, Sensor Coverage-Based)}
    \item PW-NBV-2D: An extension of NBV-2D from a pointwise to a pathwise approach. The robot performs raycasting from multiple points along the path on $O_t$, aggregates the sensor coverage as information gain, and normalizes it by the A* path distance. \textit{(Pathwise, No Prediction, Sensor Coverage-Based)}
    \item UPEN \cite{georgakis2022uncertainty}: A map-prediction-based approach that leverages an ensemble of predicted maps to estimate uncertainty. Information gain is computed as the cumulative sum of variances (uncertainty) along the path, normalized by path distance. \textit{(Pathwise, Prediction-Based, No Sensor Coverage, but Uncertainty-Only)}
    \item MapEx \cite{ho2024mapex}: A map-prediction-driven exploration frontiers are scored using pointwise probabilistic raycasting on predicted maps, combined with uncertainty map, and normalized by Euclidean distance. \textit{(Pointwise, Prediction-Based, Sensor Coverage and Uncertainty)}
\end{itemize}

\subsection{Metrics}
We assess the performance of the planners using the IoU (Intersection over Union) of the occupied cells in the predicted and ground truth maps. IoU measures the quality of the predicted 2D occupancy map, by comparing the map's predicted occupied cells with the ground truth occupancy, calculated as $\text{IoU} = \frac{\text{TP}}{\text{FP} + \text{FN} + \text{TP}}$. Higher IoU means the algorithm can create a more accurate predicted map. We also add a small buffer in the IoU calculations to not penalize small pixel-wise errors in the map, but rather targeting general map accuracy. Methods that leverage prediction (\PlannerName, MapEx, UPEN) have a clear definition of predicted maps, whereas methods without prediction (NBV-2D, PW-NBV-2D, Nearest) lack them. For these non-prediction-based methods, we collect the observed maps at every time step and generate predicted maps afterward to define predicted IoU.

\begin{figure}[t]
\centering
\includegraphics[width=\linewidth]{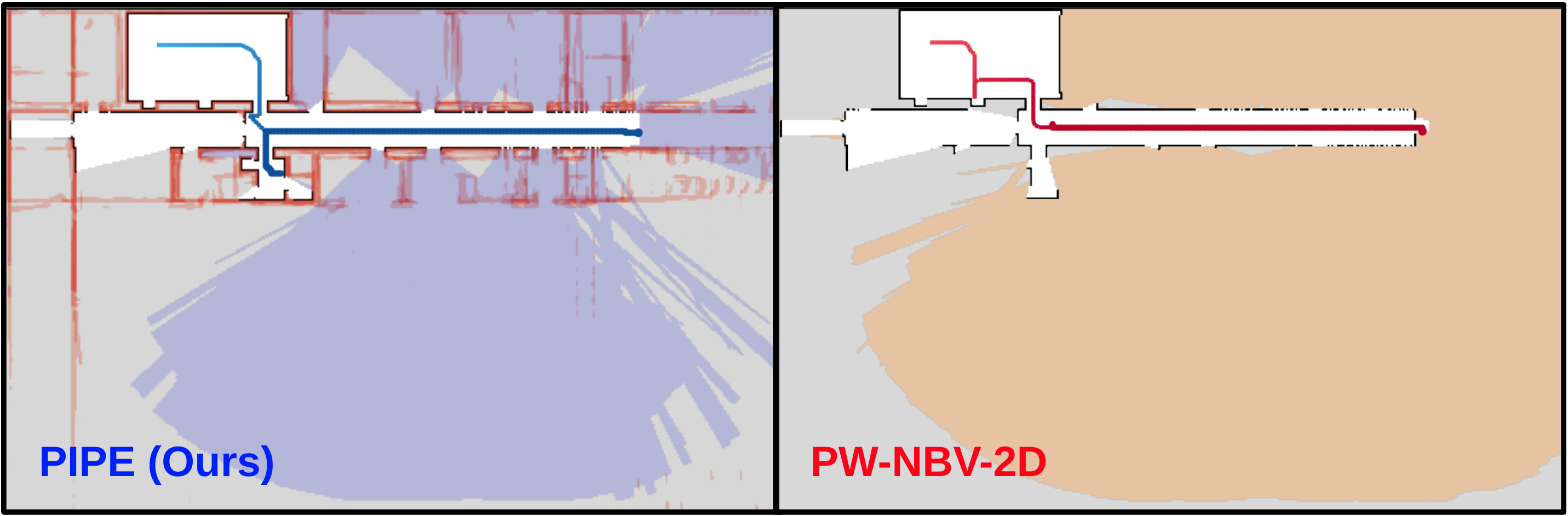}    
\caption{Qualitative comparison between sensor coverages estimated by \PlannerName~(left) and PW-NBV-2D (right). PW-NBV-2D overestimates cumulative sensor coverage, while \PlannerName~uses predictions to mitigate it.}
\label{qualitative-example}
\vspace{-0.35cm}
\end{figure}

\begin{table*}[t]
\centering
\renewcommand{\arraystretch}{1.3}
\resizebox{\textwidth}{!}{%
\begin{tabular}{
  l  %
  !{\vrule width 1.2pt} c c c c  %
  !{\vrule width 1.2pt} c c c c  %
  !{\vrule width 1.2pt} c c c c  %
}
\noalign{\hrule height 1.2pt}

\multicolumn{1}{c!{\vrule width 1.2pt}}{}%
& \multicolumn{4}{c!{\vrule width 1.2pt}}{\textbf{Small Maps (50 runs)}}%
& \multicolumn{4}{c!{\vrule width 1.2pt}}{\textbf{Medium Maps (75 runs)}}%
& \multicolumn{4}{c}{\textbf{Large Maps (50 runs)}}\\
\noalign{\hrule height 1.2pt}

\textbf{Method}  
& \textbf{90\% IoU} & \textbf{FR} (\%) & \textbf{95\% IoU} & \textbf{FR} (\%) 
& \textbf{90\% IoU} & \textbf{FR} (\%)& \textbf{95\% IoU} & \textbf{FR} (\%) 
& \textbf{90\% IoU} & \textbf{FR} (\%)& \textbf{95\% IoU} & \textbf{FR}(\%)  \\
\noalign{\hrule height 0.5pt}

\PlannerName~(ours) &
      $\mathbf{505\!\pm\!41}$ & 0 &
      $\mathbf{718\!\pm\!63}$ & 0 &
      $\mathbf{1039\!\pm\!56}$ & 0 &
      $\mathbf{1497\!\pm\!100}$ & 0 &
      $\mathbf{1979\!\pm\!115}$ & 0 &
      $\mathbf{2658\!\pm\!103}$ & 0 \\

    MapEx\,\cite{ho2024mapex} &
      $610\!\pm\!38$ & 0 &
      $752\!\pm\!58$ & 0 &
      $1165\!\pm\!72$ & 0 &
      $1602\!\pm\!101$ & 3 &
      $2266\!\pm\!164$ & 0 &
      $3399\!\pm\!217$ & 2 \\

    PW\textendash NBV\textendash 2D &
      $519\!\pm\!40$ & 0 &
      $784\!\pm\!94$ & 12 &
      $1643\!\pm\!146$ & 15 &
      $1956\!\pm\!178$ & 51 &
      $3732\!\pm\!243$ & 4 &
      $4825\!\pm\!280$ & 32 \\

    NBV\textendash 2D\,\cite{visser2008beyond} &
      $577\!\pm\!34$ & 0 &
      $852\!\pm\!70$ & 0 &
      $1321\!\pm\!65$ & 0 &
      $1694\!\pm\!104$ & 4 &
      $2336\!\pm\!89$ & 0 &
      $3037\!\pm\!116$ & 0 \\

    UPEN\,\cite{georgakis2022uncertainty} &
      $833\!\pm\!85$ & 34 &
      $1009\!\pm\!107$ & 64 &
      $2171\!\pm\!168$ & 40 &
      $2337\!\pm\!175$ & 63 &
      $4085\!\pm\!301$ & 18 &
      $4875\!\pm\!394$ & 62 \\

    Nearest\,\cite{yamauchi1997frontier} &
      $1050\!\pm\!57$ & 8 &
      $1160\!\pm\!51$ & 12 &
      $2273\!\pm\!77$ & 11 &
      $2524\!\pm\!81$ & 33 &
      $4109\!\pm\!249$ & 2 &
      $4634\!\pm\!199$ & 4 \\
    \bottomrule
  \end{tabular}
}
\caption{%
Comparison of the time–steps required to reach 90\,\% and 95\,\% IoU on ground–truth maps.
Values are reported as $\textit{mean}\,\pm\,$\emph{95\,\% confidence interval};
failed runs are excluded when computing the means and CIs.
The failure rate (FR) column shows the percentage of runs that did not attain the
target IoU within the maximum time‑steps (1,500 for \textbf{small},
3,000 for \textbf{medium}, and 6,000 for \textbf{large} maps).
Lower numbers indicate better performance.}
\label{tab:Complete_Search}
\end{table*}

\begin{figure*}[t]
\centering
\includegraphics[trim={1.0cm 6.5cm 2.7cm 1.3cm},clip,width=\linewidth]{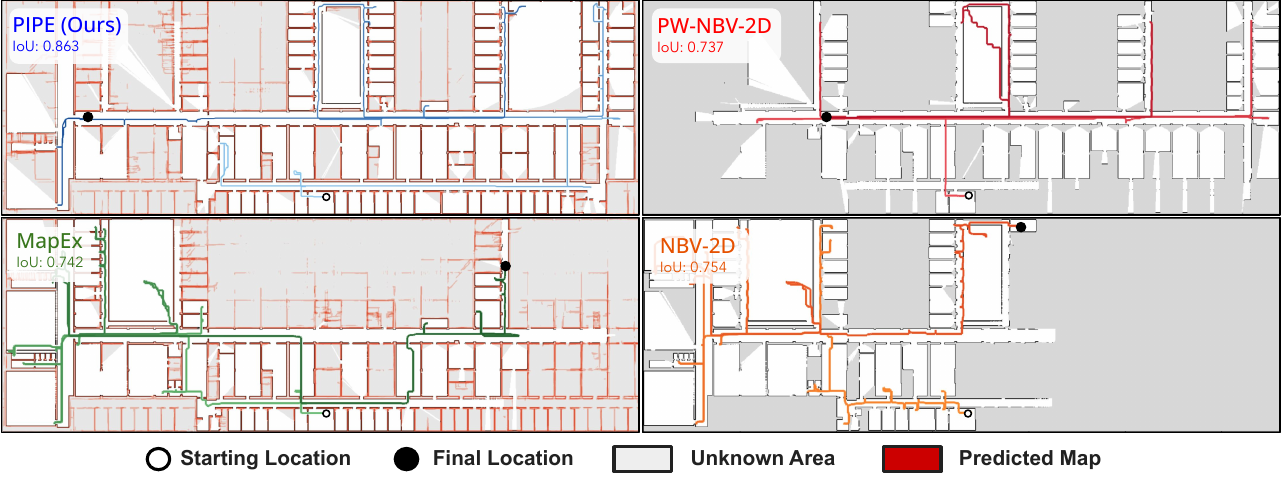}    
\caption{Qualitative comparison of methods at time step $T = 2000$ on Map 6, with the path transitioning from light to dark to indicate the passage of time. \PlannerName~performs the best, with more even exploration across the space and less backtracking compared to other methods.  }
\label{trajectory-comparison}
\vspace{-0.25cm}
\end{figure*}

\subsection{Experimental Results}
In this section, we first present the quantitative evaluation of our experiments. We analyze how much each method explores within a fixed time budget in Sec.~\ref{IoU Comparison} and examine the time required to complete map exploration, along with their later-stage behaviors, in Sec.~\ref{Complete Map Exploration}.

\subsubsection{IoU Comparison with time budgets}
\label{IoU Comparison}

Fig.~\ref{fig:result} compares exploration progress over time across three map scales for each method. Overall, we observe that \PlannerName~consistently outperforms all other methods, demonstrating stable and high performance across all map sizes. In small maps, \PlannerName~achieves the highest final IoU, surpassing PW-NBV-2D by 1.15\%, MapEx by 7.08\%, NBV-2D by 4.13\%, UPEN by 17.91\%, and Nearest by 26.99\%.

In the medium-sized maps, \PlannerName~maintains its lead, though the ranking among the baselines shifts. Here, MapEx emerges as the second-best method, outperforming the rest. At $T=1000$, \PlannerName~exceeds PW-NBV-2D by 11.69\%, MapEx by 4.93\%, NBV-2D by 8.99\%, UPEN by 22.89\%, and Nearest by 23.45\%. 

In large maps, \PlannerName, MapEx, and NBV-2D perform at comparable levels, but PW-NBV-2D experiences a significant drop in performance. At $T=2000$, \PlannerName~outperforms MapEx by 2.1\%, NBV-2D by 4.73\%, UPEN by 16.85\%, and Nearest by 22.57\%. Although the final IoU differences between the top three methods appear moderate, the trajectory suggests that MapEx is reaching saturation, while \PlannerName’s IoU continues to increase. This trend is further analyzed in Sec.~\ref{Complete Map Exploration}. Additionally, \PlannerName~surpasses PW-NBV-2D by 15.20\% in large maps---a notably larger gap than in small and medium maps---indicating that pathwise integration of information gain alone is insufficient and can even degrade performance when combined with certain estimation methods.

The performance of each method can also be measured with Area Under the Curve (AUC) of each method's plot. According to Table \ref{AUC}, \PlannerName~surpassed all other methods for every map size, verifying its efficient map exploration.

\subsubsection{Complete Map Exploration}
\label{Complete Map Exploration}
In addition to evaluating each method's performance under fixed time budgets, we also analyzed their ability to achieve complete map exploration. Here, we define this as reaching at least 95\% IoU with the ground truth map. To better understand the long-tail behavior of the methods, we also report results for 90\% IoU, highlighting the difficulty of achieving the additional 5\% increase. For both IoU thresholds, we measure the number of time steps required by each method.

The results are shown in the Table~\ref{tab:Complete_Search}. Overall, \PlannerName~outperformed all other methods in both efficiency and stability. It required the fewest time steps to reach both 90\% and 95\% IoU across all maps and was the only method to achieve a zero failure rate for complete exploration at both thresholds.

\subsubsection{Qualitative Discussion}
As shown in Fig.~\ref{fig:result} and Table~\ref{AUC}, pathwise methods (\PlannerName, PW-NBV-2D) outperform pointwise methods (MapEx, NBV-2D) in small maps. However, in medium and large maps, \PlannerName~continues to outperform MapEx, while PW-NBV-2D performs notably worse. This suggests that simply integrating cumulative sensor coverage can degrade performance as map size increases. The qualitative comparison in Fig.~\ref{qualitative-example} provides insights into why PW-NBV-2D struggles in larger maps. \PlannerName~computes sensor coverage using probabilistic raycasting over predicted maps, providing a more conservative and accurate estimation of unobserved areas. In contrast, PW-NBV-2D significantly overestimates sensor coverage in unobserved regions. While the accumulation of errors along the pathwise raycasts may have little noticeable impact in small maps, it becomes increasingly significant as map size and complexity increase.

Lastly, Fig.~\ref{trajectory-comparison} presents an example of the exploration progress of \PlannerName, MapEx, PW-NBV-2D, and NBV-2D. All methods start from the same location. The two pointwise methods, NBV-2D and MapEx, exhibit similar behaviors, primarily focusing on the left half of the map. This highlights a key limitation of pointwise methods—their tendency to fixate on short-sighted local information gain. Meanwhile, PW-NBV-2D initially explores effectively but later gravitates toward traversing long corridors, skipping rooms. In contrast, \PlannerName~leverages both predictive uncertainties and pathwise coverage, balancing exploration between rooms and corridors, ultimately achieving the best performance.

\section{Conclusion}
\label{sec:conclusion}
In this work, we presented \PlannerName, a planner based on Pathwise Information Gain with Map Prediction for Exploration. Our approach addresses the need to compute information gain cumulatively along a path rather than merely at a point. To achieve this efficiently, we proposed techniques to reduce computational overhead and incorporated map predictions to mitigate the overestimation of cumulative sensor coverage. Experiments on a real-world floorplan dataset simulator demonstrated that \PlannerName~outperforms baselines, including map-prediction-based exploration method relying solely on pointwise gain and pathwise exploration methods without map prediction. For future work, we aim to extend our research to real-world robot deployment.

\balance

\bibliographystyle{ieeetr}
\bibliography{references}

\end{document}